\documentclass[10pt, a4paper]{article}

\usepackage{lrec-coling2024} 

\usepackage{natbib}
\usepackage{boxedminipage}
\usepackage{multibib}
\makeatletter
\def\@mb@citenamelist{cite,citep,citet,citealp,citealt,citepalias,citetalias}
\makeatother
\newcites{languageresource}{~}

\usepackage{graphicx}
\usepackage{tabularx}
\usepackage{soul}
\usepackage{enumitem}

\usepackage{adjustbox}
\usepackage{url}
\usepackage[utf8]{inputenc}
\usepackage[T2A,T1]{fontenc}
\usepackage{booktabs}
\usepackage{color}
\usepackage{multirow}

\newcommand\textcyr[1]{{\fontencoding{T2A}\selectfont #1}}
\newcommand\comment[1]{}

\usepackage{titlesec}
\titleformat{\section}{\normalfont\large\bfseries\center}{\thesection.}{1em}{}
\titleformat{\subsection}{\normalfont\SmallTitleFont\bfseries\raggedright}{\thesubsection.}{1em}{}
\titleformat{\subsubsection}{\normalfont\normalsize\bfseries\raggedright}{\thesubsubsection.}{1em}{}
\renewcommand\thesection{\arabic{section}}
\renewcommand\thesubsection{\thesection.\arabic{subsection}}
\renewcommand\thesubsubsection{\thesubsection.\arabic{subsubsection}}

\usepackage{xcolor}
\usepackage{hyperref}
 \definecolor{darkblue}{rgb}{0, 0, 0.5}
  \hypersetup{colorlinks=true, citecolor=darkblue, linkcolor=darkblue, urlcolor=darkblue}

\usepackage{xstring}

\usepackage{color}

\title{Cross-lingual Named Entity Corpus for Slavic Languages}

\name{Jakub Piskorski,$^{\star}$ Michał Marcińczuk,$^{\spadesuit}$ Roman Yangarber$^{\dagger}$} 

\address{$^{\star}$Polish Academy of Sciences, Poland \\
        $^{\spadesuit}$Samurai Labs, Poland \\
        $^{\dagger}$University of Helsinki, Finland \\
         jpiskorski@gmail.com, marcinczuk@gmail.com, roman.yangarber@helsinki.fi}

\abstract{
This paper presents a corpus manually annotated with named entities for six Slavic languages---Bulgarian, Czech, Polish, Slovenian, Russian, and Ukrainian.  This work is the result of a series of shared tasks, conducted in 2017--2023 as a part of the Workshops on Slavic Natural Language Processing.  The corpus consists of 5\,017 documents on seven topics. 
The documents are annotated with five classes of named entities.  
Each entity is described by a category, a lemma, and a unique cross-lingual identifier.  We provide two train-tune dataset splits---\textit{single topic out} and \textit{cross topics}.  
For each split, we set benchmarks using a transformer-based neural network architecture with the pre-trained multilingual models --- \textit{XLM-RoBERTa-large} for named entity mention recognition and categorization, and \textit{mT5-large} for named entity lemmatization and linking. \\ 
 \\ \newline  }

\begin{document}

\maketitleabstract

\section{Introduction}

High-quality recognition and analysis of named entities (NEs) is essential for many information access tasks, such as document retrieval and clustering.  It also constitutes a fundamental step in a wide range of natural language processing (NLP) pipelines for higher-level analysis of text, such as information extraction~\cite{grishman-2019-25-years-IE,huttunen:2002-lrec-IE-diversity}, and anonymization of sensitive data~\cite{10.1007/s10207-022-00607-5}.

In this paper, we present a text corpus manually annotated with named entities for six Slavic languages---Bulgarian, Czech, Polish, Slovenian, Russian, and Ukrainian. This corpus results from a series of shared tasks on Named-Entity Recognition, Normalization and Linking~\cite{piskorski-etal-2019-second,piskorski-etal-2021-slav,yangarber-etal-2023-slav}, conducted in 2017--2023 as a part of Workshops on Balto-Slavic Natural Language Processing,\footnote{\url{https://bsnlp.cs.helsinki.fi/}} whose goal is to stimulate research and foster the creation of tools and resources for these languages, which have over 400 million native speakers.

The corpus consists of 5~017 documents, mainly online news articles, covering seven topics highly debated in the news. The documents are annotated with five categories of named entities: person, organization, location, named mentions of events, and product names---where the last category covers artefacts typically mentioned in the news, e.g., services, awards, cultural artefacts, broadcast media programs, newspapers, legal acts, etc.  Each mention of an entity is described with a category, a lemma, a cross-language identifier and positional information. In total, 152 888 named-entity mentions have been annotated, making this the largest cross-lingual NE corpus for Slavic languages. 

We also provide two train-tune dataset splits---single-topic-out and cross-topic. For each split, we set benchmarks using a transformer-based neural network architecture with the pre-trained multilingual models --- \textit{XLM-RoBERTa-large} for named entity mention recognition and categorization, and \textit{mT5-large} for named entity lemmatization and linking.

The rationale behind the creation and release of the presented corpus is manifold. First, it is a unique new resource in: (a) covering multiple Slavic languages, and documents revolving around the {\em same topics} across languages, (b) {\em linking} the names  cross-lingually, and (c) providing {\em base forms} of names. Other existing NER corpora for Slavic languages are strictly monolingual. Secondly, the dataset results from 4 editions of shared tasks on NER for Slavic languages, which attracted a large amount of participants, and is an integration and curation of all datasets used in the tasks with comprehensive annotation guidelines.  
We have receive inquiries and requests from the research community on the availability of the corpus for training NER, NEL and lemmatisation models that goes beyond the shared task itself.  For the first time the integrated and curated corpus is available with positional anchoring---required by many of the approaches applied to the aforementioned shared tasks.

We believe the corpus contributes to fostering research not only on NER, but also name linking and name lemmatization for Slavic languages.  Slavic languages exhibit certain challenging phenomena, like inflection (up to 7 nominal cases), e.g., in newspapers, a significant amount of NE occurrences (i.e., 30-50\%) are not nominative forms, and computing base forms of names is a complex task~\cite{journals/ir/PiskorskiWS09}.  The corpus covers the domain of online news, which is another factor making it attractive for research on news analysis. 

The paper is organized as follows.  In Section~\ref{sec:related} we present related work.
In Section~\ref{sec:annotations}, we describe the process of creating the corpus, including the taxonomy, annotation process, and corpus statistics.
In Section~\ref{sec:models}, we present the results of evaluation of various benchmark models on the named entity recognition (NER), name lemmatization, and entity linking tasks.
We end with conclusions and outlook on future research in Section~\ref{sec:conclusions}.

\section{Related Work}
\label{sec:related}

The task of NE recognition has been studied since the 1990's, with a vast amount of research on various approaches, ranging from knowledge-based to machine-learning~\cite{nouvel2016named,yadav-bethard-2018-survey,JEHANGIR2023100017}.
The research on NER was initally fostered by the NER-related shared tasks in the context of the Message Understanding Conferences (MUCs)~\cite{chinchor:98} and the ACE Programme~\cite{conf/lrec/DoddingtonMPRSW04}.
The first \emph{multilingual} NER shared task, which covered several European languages, including Spanish, German, and Dutch, was organized in the context of the the CoNLL conferences~\cite{TjongKimSang:2002:ICS:1118853.1118877,TjongKimSang:2003:ICS:1119176.1119195}.
The NE types covered in these campaigns covered mainly the ``standard'' types: {\em person, organisation}, and {\em location}.
The task of Entity Discovery and Linking
(EDL)~\cite{ji:ea:2014,ji:ea:2015} emerged as a track of the NIST Text Analysis Conferences (TAC).
EDL aims to extract entity mentions from a collection of documents in multiple languages
(English, Chinese, and Spanish), and to partition the entities into cross-document
equivalence classes, by either linking mentions to a knowledge base or directly clustering them.

Related to cross-lingual NE recognition is NE transliteration, i.e., linking NEs across languages that use different alphabets/writing systems.  
A series of NE Transliteration Shared Tasks were organized as part of NEWS---Named Entity Workshops \cite{duan2016report}, focusing mostly on Indian and Asian languages.  In 2010, the NEWS Workshop included a shared task on Transliteration Mining~\cite{kumaran2010report}, i.e., mining of names from parallel corpora, in English, Chinese, Tamil, Russian, and Arabic.


Research on NER focusing on Slavic languages includes NER for Croatian \cite{karan2013croner,ljubesic2013combining}; NER in Croatian tweets \cite{baksa2017tagging}; a manually annotated NE corpus for Croatian \cite{agic2014setimes}; NER in Slovene~\cite{stajner2013razpoznavanje,ljubesic2013combining}; a Czech corpus of 11K annotated NEs~\cite{vsevvcikova2007named}; NER for Czech~\cite{Konkol:2013}; tools and resources for fine-grained annotation of NEs in the National Corpus of Polish~\cite{was:etal:10,sav:pis:11}; lemmatization of NEs for Polish~\cite{journals/ir/PiskorskiWS09,marcinczuk-2017-lemmatization}.
Shared tasks on NER for Polish were organized under the umbrella of POLEVAL\footnote{\url{http://poleval.pl}}~\cite{ogr:kob:18:poleval,ogr:kob:20:poleval} and LESZCZE\footnote{\url{https:/lepiszcze.ml/tasks/namedentityrecognition}} campaigns. 
Recent shared tasks on NER in Russian include~\cite{alexeeva2016factrueval,df1825c3ce5047c094b8dce8f40a6861}, with the latter based on NEREL---a Russian dataset for NER and relation extraction, described in \citealp{loukachevitch-etal-2021-nerel}.
{SemEval 2022 included Task 11: MultiCoNER Multilingual Complex Named Entity Recognition}\footnote{\url{https://multiconer.github.io/multiconer\_1}}~\cite{malmasi-etal-2022-semeval} and {SemEval 2023 included Task 2: MultiCoNER II Multilingual Complex Named Entity Recognition},\footnote{\url{https://multiconer.github.io}}~\cite{fetahu-etal-2023-semeval} which included Russian and Ukrainian respectively.

A series of Shared Tasks on multilingual NE recognition, normalization and cross-lingual matching for Slavic languages have been organized in the context of the ACL-sponsored workshops on NLP for Slavic languages,\footnote{\url{https://bsnlp.cs.helsinki.fi/}}~\cite{piskorski-2017-first-BSNLP-challenge,piskorski-etal-2019-second,piskorski-etal-2021-slav,yangarber-etal-2023-slav} which were the first attempts at such shared tasks covering multiple Slavic languages.
The corpus we present in this paper results from merging and curating the corpora from the second (2019), third (2021), and fourth editions (2023) of these shared tasks, where, in contrast to the original datasets, the name mentions are associated with positional information.

\section{Named-Entity Annotation}
\label{sec:annotations}

This section describes the end-to-end NE annotation process, including: NE taxonomy, document acquisition, document annotation, the annotation format and the resulting corpus statistics.

\subsection{Taxonomy}

The Named Entity taxonomy consists of five NE classes: person, organization, location, event and product.
The choice of these five entity types has been mainly motivated by the the domain of the corpus, namely, online news centred around certain highly-debated topics. In particular, this applies to the inclusion of the less standard entity types, i.e., events (as they are referred to in the news) and products, which cover named mentions of artefacts typically mentioned in the news, e.g., product names, services, awards, cultural artefacts, newspapers, legal acts, etc.

Apart from the entity type, entity annotation also includes information on the base form of the entity, and a cross-lingual ID, so that mentions of the same real-world entity across multiple documents can be assigned to the same unique ID.  

More detailed definitions of the five entity categories are provided below.

\begin{description}[style=unboxed,leftmargin=0.15cm]

\item[Person names (\textsc{PER}):] 

Names of real or fictional persons, including initials and pseudonyms, but without including titles, honorifics, and functions or positions.  Toponym-based named references to groups of people that have no formal organization unifying them (e.g., {\color{blue}``\textit{Ukrainians}''}) 
and named mentions of other groups of people that do have a formal organization unifying them (e.g., {\color{blue}``\textit{The Royals won}''}) should be both tagged as \textsc{PER}. Finally, also personal possessives derived from a person's name should be annotated as \textsc{PER}.
  



\item[Locations (\textsc{LOC}):]

  All toponyms and geopolitical entities---cities, counties, provinces, countries,
  regions, bodies of water, land formations, etc.---including named mentions of {\em
    facilities}---e.g., stadiums, parks, museums, theaters, hotels, hospitals,
  transportation hubs, churches, streets, railroads, bridges, and similar facilities. Named mentions of facilities {\em also} refering to an organization should be tagged with \textsc{LOC}, in {\color{blue}``\textit{The Franciszek Raszeia Hospital hired new staff due to the covid pandemics.}}'' the hospital mention should be annotated as \textsc{LOC}.


\item[Organizations (\textsc{ORG}):]
 All organizations, including companies, public institutions, political parties, international, religious, sport, educational and research organizations, etc. Organization designators and mentions of the seat of the organization are considered to be part of the organization name. 
  

\item[Products (\textsc{PRO}):]

  All names of {\em products and services}, such as electronics, cars, newspapers,  web-services, medicines, awards, books, media programmes, initiatives, legal documents, and treaties, e.g., {\color{blue}``\textit{Maastricht Agreement}''}, etc.

\item[Events (\textsc{EVT}):] named mentions of incidents, occasions, and events that refer to a specific point of time or a time span, including conferences, concerts, festivals, holidays, e.g., {\color{blue}``\textit{Christmas 2024}''}, wars, battles, disasters, e.g., {\color{blue}``\textit{1943 Gibraltar Liberator AL523 crash}''}, outbreaks of infectious diseases e.g., {\color{blue}``\textit{Covid-19}''}, and also speculative, and fictive events.


  
\end{description}

In case of complex {\em nested} named entities, only the {\em top-most} entity is annotated.

In case of coordinated phrases mentioning entities, both elements are to be annotated separately, e.g., in {\color{blue}``\textit{French and German Ministry of Foreign Affairs},''} both {\color{blue}``\textit{French}''} and
{\color{blue}``\textit{German Ministry of Foreign Affairs},''} are to be annotated.

Detailed Annotation Guidelines are provided in Annex~\ref{ne-annotation-guidelines}.

\subsection{Document Acquisition}

For the creation of the corpus various topics were selected, including, the {\sc Covid-19} pandemic, the 2020 USA Presidential elections ({\sc USA 2020 Elections}),  {\sc Asia Bibi}, which relates to a Pakistani woman involved in a blasphemy case, {\sc Brexit}, {\sc Ryanair}, which faced a massive strike, {\sc Nord Stream}, the controversial Russian-European project, and the {\sc Russia-Ukraine war}.

\begin{figure}[h]
    \centering
    \includegraphics[scale=.83]{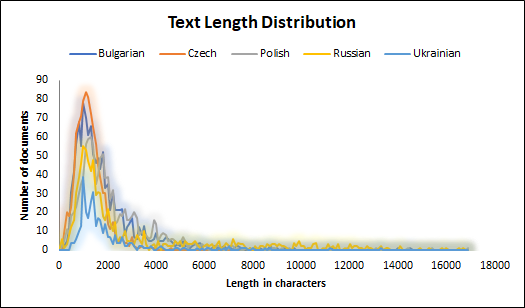}
    \caption{Distribution of the text length for all languages.}
    \label{fig:len-distro}
\end{figure}

For each of the aforementioned topics relevant documents were collected in the following manner. A search query was posed to Google and/or publicly available crawled data repositories, in each of the target languages. The query returned documents in the target language. We removed duplicates, downloaded the HTML---mainly news articles---and converted them into plain text. Since the result of HTML parsing may include
not only the main text of a Web page, but also spurious text, some additional manual cleaning was applied when necessary. The resulting set of ``cleaned'' documents were used to manually select documents for each language and topic for the final datasets. Figure~\ref{fig:len-distro} provides the text length histograms for all languages. More than 95\% of the texts are not longer than approx. 3K characters.

\subsection{Annotation Platform}

Documents were annotated using the Inforex\footnote{\url{github.com/CLARIN-PL/Inforex}} web-based platform for annotation of text corpora~\cite{DBLP:conf/ranlp/MarcinczukOK17}. The platform allows for sharing a common list of entities, and perform entity-linking semi-automatically: for a given entity, an annotator sees a list of entities of the same type inserted by all annotators and can select an entity ID from the list. 
Furthermore, it keeps track of all lemmas and IDs inserted for each surface form, and inserts them automatically to speed up the annotation process which in most cases boils down to a confirmation of the proposed values by an annotator. 
All annotations were made by native speakers.  After annotation, we performed {\em multiple phases} of automatic and manual consistency checks, to reduce annotation errors. 
The entire process is described in more detail in the next Section.

\begin{figure}[ht]
  \centering
  \includegraphics[trim= 0mm 0mm 0mm 0mm, clip, width=0.5\textwidth]{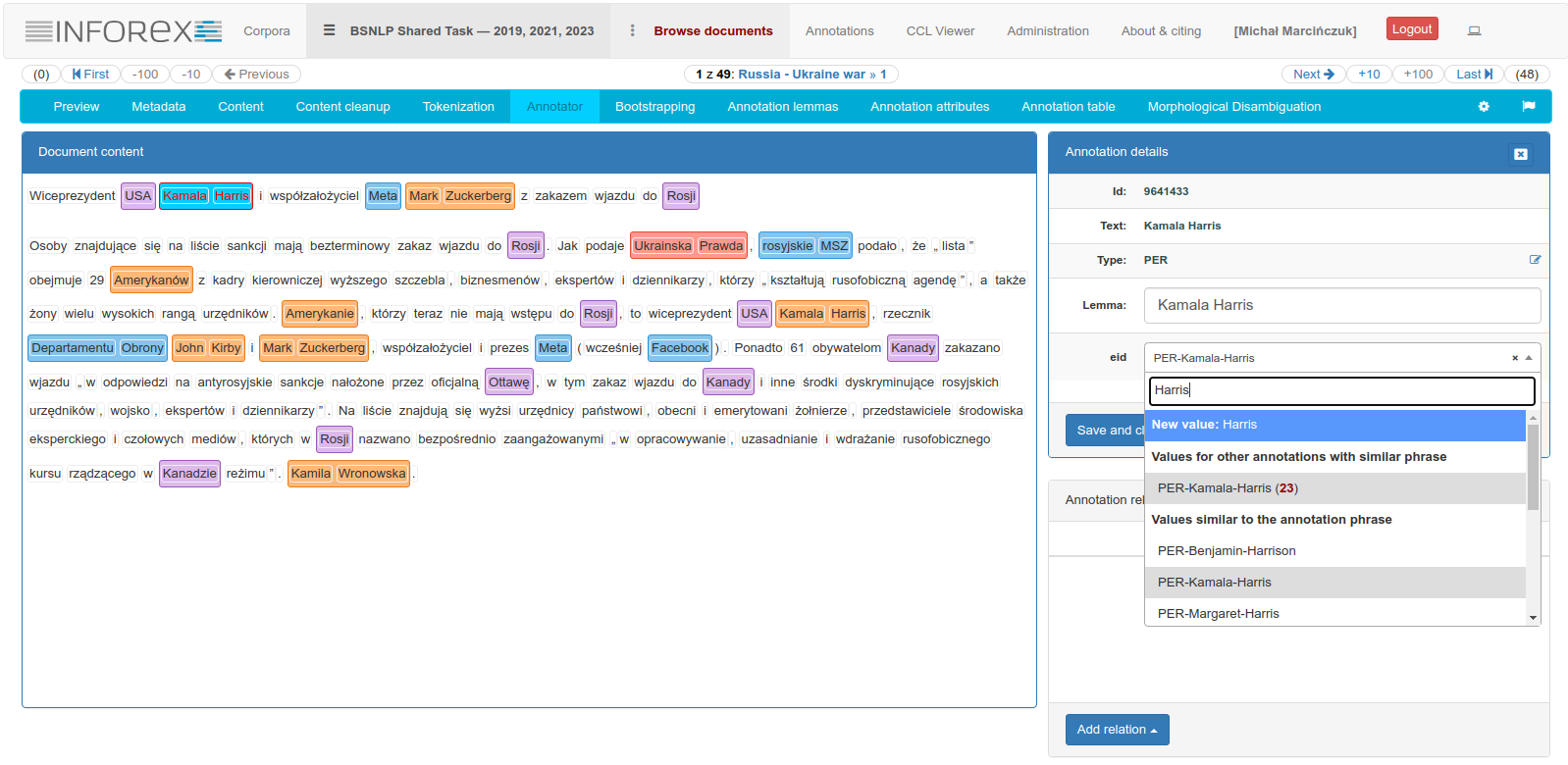}
  \caption{Screenshot of the Inforex Web interface, the tool used for data annotation.}
  \label{fig:inforex}
\end{figure}


\subsection{Annotation Process}


The annotation process includes the following steps:

\begin{enumerate}

\item all annotators were first provided with detailed annotation guidelines and Q/A sessions were organized,

\item each document was initially annotated by at least one experienced annotator, and a revision was undertaken by master annotator (most experienced annotator for the language) for a given language, 

\item to assure high quality annotations, the annotation platform, namely Inforex, was extended with features to automatically suggest the base form and cross-lingual ID for the candidate entities based on all accumulated annotations over time, which turned beneficial for the human annotators,

\item finally, several iterations of data curation (across languages) were carried out with the annotators to resolve any potential inconsistencies (computed automatically), e.g.:

\begin{itemize} 
\item same or similar surface forms were assigned different NE types in different documents, 
\item same surface forms were assigned different base forms, 
\item same surface form were assigned different cross-lingual IDs. 
\end{itemize}

\noindent In the last case, it is worth mentioning that there was an agreed upon and well-defined convention of how cross-lingual IDs are created in order to avoid potential errors.
\end{enumerate}

We encountered some challenges in the annotation process sketched above.

Firstly, there were some disputes regarding what falls under specific NE categories and what does not; this was resolved through various iterations with annotators who annotated texts in different languages, and allowed to converge to a final version of the annotation guidelines.

Secondly, there were disagreements related, in particular, to the ambiguities between \textsc{ORG} vs. \textsc{PER}, and \textsc{PRO} vs. \textsc{ORG} in terms of deciding on the NE type.  
Here, specific disambiguation rules were introduced in the guidelines, e.g., in cases, in which the local document context and common knowledge does not provide sufficient information to disambiguate the named entity type (e.g., in the phrase {\color{blue}\textit{Opel announced that ...}} the mention of {\color{blue}\textit{Opel}} could potentially refer either to an organisation (ORG) or a person (PER), namely, \textit{Adam Opel}, the founder of the company, the more probable interpretation should be considered, i.e., ORG, since Adam Opel died in 1837 and could not announce anything recently, unless the document is a historical one (unlikely). In case both NE type interpretations appear to be equally probable NE type disambiguation rules should be applied. In this particular case, if possible interpretations are ORG or PER, then PER should have priority.

The aforementioned NE type ambiguity problems had a direct impact on assigning the correct cross-lingual IDs. 

\subsection{Annotation Format}

For each text document in the dataset a corresponding
annotation file exists, which includes for each NE mention
a line in the following format:

\begin{verbatim}
<START> <END> <MENTION> <BASE> <CAT> <ID>
\end{verbatim}

\noindent where \verb+<START>+, \verb+<END>+ are the positional information, \verb+<MENTION>+ is the surface form of the mention,
\verb+<BASE>+ is the base form of the entity, \verb+<CAT>+ is the 
the category of the entity (ORG, PER, LOC, PRO, or EVT), and \verb+<ID>+ is the cross-lingual NE identifier. All the elements are separated by tabs. An example of a raw document in Polish and a corresponding annotation file is shown in Figure~\ref{fig:format}.

\begin{figure}
\begin{center}
\begin{boxedminipage}{1.0\linewidth}
\begin{scriptsize}

\textbf{Input:}\\[1em]
{\color{blue}
\noindent{\underline{Tusk} o \underline{Brexicie}: Nie jest za późno. Nasze serca są otwarte:

Nasze serca wciąż są dla was otwarte - powiedział \underline{Donald Tusk} kilka dni po tym, 
jak opublikowano sondaż mówiący, że kolejne referendum ws. \underline{Brexitu} mogłoby pokazać, 
że \underline{Brytyjczycy} wolą zostać w \underline{Unii Europejskiej}.

Nie jest za późno, aby powstrzymać \underline{Brexit}, a \underline{Wielka Brytania} wciąż może zmienić zdanie 
- powiedział przewodniczący \underline{Rady Europejskiej} eurodeputowanym w \underline{Strasburgu}. --- Jeśli rząd \underline{Wielkiej Brytanii} 
będzie trzymał się swojej decyzji o odejściu, \underline{Brexit} stanie się rzeczywistością z wszystkimi jej negatywnymi konsekwencjami. 
Czyż sam \underline{David Davis} nie powiedział: “jeśli demokracja nie może zmienić zdania, przestanie być demokracją”? --- dodał. }}
{\ }{\ }\\\\
\textbf{Output:}
\begin{verbatim}
0 3 Tusk Tusk PER PER-Donald-Tusk
5 12 Brexicie Brexit EVT EVT-Brexit
90 99 Donald Tusk Donald Tusk PER PER-Donald-Tusk
165 171 Brexitu Brexit EVT EVT-Brexit
189 199 Brytyjczycy Brytyjczycy PER GPE-Great-Britain
211 226 Unii Europejskiej Unia Europejska ORG ORG-EU
257 262 Brexit Brexit EVT EVT-Brexit
265 278 Wielka Brytania Wielka Brytania LOC GPE-UK
326 341 Rady Europejskiej Rada Europejska ORG ORG-EU-Counc
358 367 Strasburgu Strasburg LOC GPE-Strasbourg
379 394 Wielkiej Brytanii Wielka Brytania LOC GPE-UK
434 439 Brexit Brexit EVT EVT-Brexit
511 520 David Davis David Davis PER PER-David-Davis
604 613 Donald Tusk Donald Tusk PER PER-Donald-Tusk
\end{verbatim}
\end{scriptsize}
\end{boxedminipage}
\caption{An example of a raw document in Polish and a corresponding NE-annotated file.}
\label{fig:format}   
\end{center}
\end{figure}

\subsection{Statistics}

\begin{figure}[h]
    \centering
    \includegraphics[scale=.61]{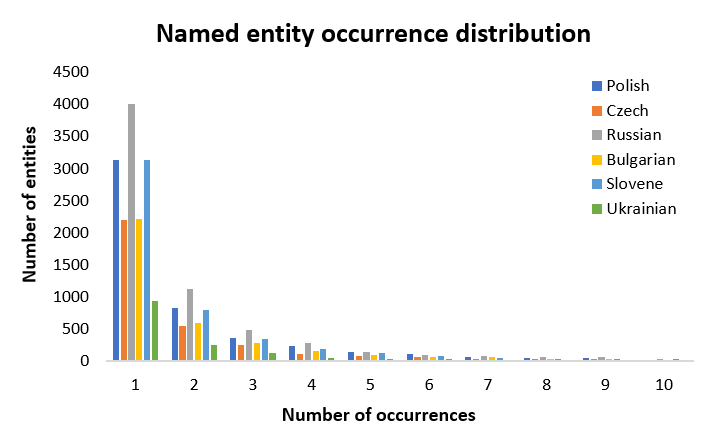}
    \caption{Distribution of named-entity occurrences for all languages.}
    \label{fig:ne-oc-distro}
\end{figure}

High-level statistics for the entire corpus in terms of the number of documents and text spans annotated are provided in Table~\ref{tab:overall-stats}. The corpus consists of 5017 documents with 152 888 NE mentions in six languages: Polish, Czech, Russian, Bulgarian, Slovene, and Ukrainian. 
Table~\ref{tab:type-breakdown} provides a comparison of the distribution of NE types across languages. Approximately 80\% of the entities are person, organization, or location names. There are, in total, 25~216 unique NE text forms. Of these, 14 831 forms are {\em hapax legomena} — these forms appears only once in the entire dataset  (9.7\% of all entities and 58.8\% of unique entities). 194 NE forms (approx. 0.8\% of all forms) appear at least 100 times in the entire corpus. Figure~\ref{fig:ne-oc-distro} shows the distribution of NE occurrence for all languages, whereas Table~\ref{tab:most-freq-stats} provides the 5 most frequent NE mentions for each language.


The detailed statistics on the distribution of named-entity types, unique surface forms, lemmas, and entity IDs per topic (subcorpus) are provided in Annex~\ref{sec:stats}.

\begin{table*}[ht]
  \centering
  \begin{scriptsize}
    \begin{tabular}{@{} lrrrrrrrrrrrr|r}
      \toprule
      & \multicolumn{2}{c}{\sc PL} & \multicolumn{2}{c}{\sc CS} & \multicolumn{2}{c}{\sc RU} & \multicolumn{2}{c}{\sc BG} & \multicolumn{2}{c}{\sc SL} & \multicolumn{2}{c}{\sc UK} & \multicolumn{1}{c}{\sc Total} \\      
 topic & \#doc & \#an & \#doc & \#an & \#doc & \#an & \#doc & \#an & \#doc & \#an & \#doc & \#an & \#doc \\ 
 \midrule
{\sc Brexit}             & 500 & 16~914 & 284 & 5~706 & 153 &  4~481 & 600 & 16~894 &  52 & 2~287 & 50 & 1~473 & 1~639 \\
{\sc Asia Bibi}          & 88 &   1~946 &  89 & 1~552 & 118 &  2~294 & 101 &  2~042 &  4  &    89 &  6 &   157 &   406 \\
{\sc Nord Stream}        & 151 &  5~032 & 161 & 4~129 & 150 &  4~210 & 130 &  3~387 &  74 & 4~185 & 40 & 1~544 &   706 \\
{\sc Ryanair}            & 146 &  2~386 & 163 & 2~502 & 150 &  2~306 &  87 &  1~176 &  52 & 1~558 & 63 & 1~194 &   661 \\
{\sc Covid-19}           & 103 &  2~272 & 155 & 2~257 &  83 &  4~907 & 151 &  3~680 & 178 & 5~503 & 85 & 2~389 &   755 \\
{\sc US 2020 Elections}  & 66 &   2~550 &  85 & 1~137 & 170 & 18~065 & 151 &  6~088 & 143 & 7~089 & 83 & 3~091 &   698 \\
{\sc Russia-Ukraine War} & 50 &   1~832 &  50 & 1~045 &  52 &  1~539 & - & - & - & - & - & -                    & 152 \\ 
\midrule
      Total              &  1~104 & 32~932 & 987 & 18~328 & 876 & 37~802 & 1~220 & 33~267 & 503 & 20~711 & 327 & 9~848 & 5~017\\
      \bottomrule
    \end{tabular}
    \end{scriptsize}
    \caption{High-level statistics: number of documents (\#doc) and text spans annotated (\#an) per dataset and language.}
  \label{tab:overall-stats}
\end{table*}

\begin{table}[ht]
  \centering
  \begin{scriptsize}
    \begin{tabular}{@{} lrrrrrr}
      \toprule
Type & {\sc PL} & {\sc CS} & {\sc RU} & {\sc BG} & {\sc SL} & {\sc UK}\\
\midrule
{\sc PER} & 7~257 & 4~430 & 10~555 & 8~304 & 6~831 & 1~935 \\
{\sc LOC} & 11~603 & 6~748 & 12~855 & 10~838 & 6~749 & 3~484\\
{\sc ORG} & 9~080 & 4~389 & 8~960 & 7~887 & 4~170 & 2~787 \\
{\sc PRO} & 2~640 & 1~356 & 3~025 & 1~957 & 1~251 & 371 \\
{\sc EVT} & 2~295 & 1~413 & 2~337 & 3~301 & 1~676 & 1~271 \\
      \midrule
      Total & 32~875 & 18~336 & 37~732 & 32~287 & 20~677 & 9~848\\
      \bottomrule
    \end{tabular}
    \end{scriptsize}
    \caption{Distribution of named-entity types across languages and statistics on surface forms, lemmas, and unique entities IDs.}
  \label{tab:type-breakdown}
\end{table}

\begin{table*}[ht]
  \centering
  \setlength{\tabcolsep}{4pt}
  \begin{scriptsize}
    \begin{tabular}{@{}rrrrrrrrrrrr}
      \toprule
      \multicolumn{2}{c}{\sc PL} & \multicolumn{2}{c}{\sc CS} & \multicolumn{2}{c}{\sc BG} & \multicolumn{2}{c}{\sc RU} & \multicolumn{2}{c}{\sc SL} & \multicolumn{2}{c}{\sc UK}  \\      
 form & \#num & form & \#num & form & \#num & form & \#num & form & \#num & form & \#num  \\ 
 \midrule
\textit{UE} & 1~590 & \textit{EU} & 702 & \textit{\textcyr{ЕС}} & 2~612 & \textit{\textcyr{США}} & 1~711 &  \textit{ZDA} & 784 & \textit{Ryanair} & 343 \\

\textit{Wielkiej Brytanii} & 895 &  \textit{brexitu} & 443 & \textit{\textcyr{Великобритания}} &  2~063 & \textit{\textcyr{Трампа}} &  754 &  \textit{Trump}  &  725 &  \textit{\textcyr{США}} & 317  \\

\textit{Nord Stream 2} &  790 & \textit{Ryanair} & 433 & \textit{\textcyr{Брекзит}} &  1~920 & \textit{Ryanair} &  637 &  \textit{EU} & 588 & \textit{\textcyr{ЄС}} & 273 \\

\textit{Unii Europejskiej} &  462 & \textit{Británie} & 392 & \textit{\textcyr{Тръмп}} &  678 &  \textit{\textcyr{Трамп}} &  633 &  \textit{Biden} & 324 & \textit{Brexit} & 235  \\

\textit{Brexit} &  456 & \textit{Nord Stream 2} & 257 & \textit{\textcyr{Лондон}} &  669 & \textit{\textcyr{России}} &  570 & \textit{covid-19} & 187 & \textit{\textcyr{Трамп}} & 181  \\
\bottomrule
    \end{tabular}
    \end{scriptsize}
    \caption{Top 5 most frequent named-entity forms per language.}
  \label{tab:most-freq-stats}
\end{table*}

\section{Baseline models}
\label{sec:models}

In this Section we introduce various baseline models and accompanying evaluation results for the three tasks at hand: named-entity recognition, name lemmatization, and entity linking.

\subsection{Named-Entity Recognition}
\label{sec:ner-models}

The baseline model follows the transformer-based neural network architecture presented by \citet{devlin-etal-2019-bert}. The model consists of three core component`s: a pre-trained language model, a dropout layer, and a linear layer that performs token classification. As the pre-trained model, we use the multilingual \textit{XLM-RoBERTa-large} model \cite{DBLP:journals/corr/abs-1911-02116}.\footnote{\url{https://huggingface.co/xlm-roberta-large}}  We use the following parameter settings for the training:
\\[1em]
\begin{tabular}{ll}
    Sequence length & 256 \\
    Dropout & 0.2 \\
    Epochs & 5 \\
    Learning rate & 5e-6 \\
    Optimizer & AdamW \\
    Scheduler & Liner without warmup \\
\end{tabular}
\\[1em]
We provide baselines for two variants of the dataset split:
\begin{itemize}
    \item \textbf{single topic out} -- the test subset is a collection of documents on the topic of US Elections 2020. The train subset contains the remaining six topics (Asia Bibi, Brexit, Nord Stream, Ryanair, Covid, and Russia-Ukraine War).
    \item \textbf{cross topics} -- the train and test subsets include documents from each of the seven topics. The test subset contains around 10\% of all documents. 
\end{itemize}

The splits intend to benchmark two different aspects of the models. The \textbf{single topic out} split evaluates model generality, i.e., how it will perform on new topics. The \textbf{cross topics} split evaluates model in-domain performance.

For the \textbf{single topic out} split, the baseline models achieved a micro-averaged F-score of 0.8503 across all language categories and categories. The results range from 0.3572 for EVT category to 0.9344 for PER category. The detailed results are presented in Table~\ref{tab:eval-single-topic-all}. For all languages, the results range from 0.7933 for RU to 0.9298 for CS (see Table~\ref{tab:eval-single-topic-languages}). 

\begin{table}[ht]
  \centering
  \begin{scriptsize}
    \begin{tabular}{@{} lrrrr}
      \toprule
  \textbf{Category}   & \textbf{Precision} & \textbf{Recall} & \textbf{F$_1$} & \textbf{Support} \\ \midrule
EVT           & 57.98    & 25.81    & 35.72     &  1~422 \\
LOC           & 92.28    & 88.68    & 90.45     & 12~247 \\
ORG           & 73.25    & 58.24    & 64.89     &  6~283 \\
PER           & 92.09    & 94.84    & 93.44     & 15~936 \\
PRO           & 69.59    & 63.64    & 66.48     &  2~068 \\
\midrule
micro avg     & 87.72    & 82.51    & 85.03     & 37~956 \\
macro avg     & 86.53    & 82.51    & 84.12     & 37~956 \\

      \bottomrule
    \end{tabular}
    \end{scriptsize}
    \caption{Evaluation of the \textbf{single topic out} split---summary for all languages.}
  \label{tab:eval-single-topic-all}
\end{table}

\begin{table}[ht]
  \centering
  \begin{scriptsize}
   \begin{tabular}{@{} lrrrr}
      \toprule
  \textbf{Language}   & \textbf{Precision} & \textbf{Recall} & \textbf{F$_1$} & \textbf{Support} \\ \midrule
PL  &   90.88   & 89.53   & 90.20     &  2~549\\
CS  &   93.98   & 92.00   & 92.98     &  1~137\\
RU  &   83.51   & 75.55   & 79.33     & 18~018\\ 
BG  &   93.99   & 89.52   & 91.70     &  6~085\\
SL  &   89.77   & 92.11   & 90.93     &  7~082\\
UK  &   88.65   & 77.96   & 82.96     &  3~085\\
      \bottomrule
    \end{tabular}
    \end{scriptsize}
    \caption{Evaluation of the \textbf{single topic out} split---micro average for each language separately.}
  \label{tab:eval-single-topic-languages}
\end{table}

For the \textbf{cross topics} split, the baseline models achieved a micro-averaged F-score of 92.22 across all languages and categories. The results range from 77.07 for PRO category to 96.17 for PER category. The detailed results are presented in Table~\ref{tab:eval-cross-topics-all}. The results for each language range from 89.13 for RU to 95.89 for PL (see Table~\ref{tab:eval-cross-topics-languages}). 

\begin{table}[ht]
  \centering
  \begin{scriptsize}
    \begin{tabular}{@{} lrrrr}
      \toprule
  \textbf{Category}   & \textbf{Precision} & \textbf{Recall} & \textbf{F$_1$} & \textbf{Support} \\ \midrule
EVT         &  88.78  &  90.46  &  89.61  &    1~163 \\
LOC         &  94.94  &  96.66  &  95.79  &    4~577 \\
ORG         &  86.29  &  90.69  &  88.44  &    3~513 \\
PER         &  95.63  &  96.72  &  96.17  &    3~752 \\
PRO         &  75.48  &  78.72  &  77.07  &     954 \\
\midrule
micro avg   &  91.03  &  93.43  &  92.22  &   13~959 \\
macro avg   &  91.10  &  93.43  &  92.25  &   13~959 \\
      \bottomrule
    \end{tabular}
    \end{scriptsize}
    \caption{Evaluation of the \textbf{cross topics} split---summary for all languages.}
  \label{tab:eval-cross-topics-all}
\end{table}

\begin{table}[ht]
  \centering
  \begin{scriptsize}
    \begin{tabular}{@{} lrrrr}
      \toprule
  \textbf{Language}   & \textbf{Precision} & \textbf{Recall} & \textbf{F$_1$} & \textbf{Support} \\ \midrule
PL  &  95.30  &  96.48  &  95.89    &  2~986 \\
CS  &  93.84  &  96.17  &  94.99    &  1~567 \\
RU  &  87.26  &  91.09  &  89.13    &  3~458 \\ 
BG  &  92.20  &  94.60  &  93.38    &  2~872 \\
SL  &  86.83  &  89.73  &  88.25    &  1~976\\
UK  &  92.28  &  92.36  &  92.32    &  1~100\\
      \bottomrule
    \end{tabular}
    \end{scriptsize}
    \caption{Evaluation of the \textbf{cross topics} split---micro average, for each language.}
  \label{tab:eval-cross-topics-languages}
\end{table}

The baseline results for the cross topics split are better than the single topic out split (0.9222 vs. 0.8503 of the F-score) due to a higher overlap of entities between the train and test subsets. Each topic has a set of entities that appears in almost any document on the given topic. For example, the names {\color{blue}\textit{Asia Bibi}}, {\color{blue}\textit{Bibi}}, and similar appear solely in the documents about Asia Bibi. In the single-topic out split, the entities specific to US Elections 2020 are less likely to appear in documents on other topics. 

Although the presented corpus originates directly from the three editions of the Shared Tasks on Multilingual Named Entity Recognition, Normalization and cross-lingual Matching for Slavic Languages~\cite{piskorski-etal-2019-second,piskorski-etal-2021-slav,yangarber-etal-2023-slav} a direct comparison of the baseline model results vis-a-vis the results obtained in the aforementioned shared tasks is not straightforward and indicative only since the task has been defined in a slightly different manner, i.e., it did not focus on the detection of positional information in the text, but on the extraction of unique named entity forms from a given document, where multiple occurrences of the same form are counted only once. 

Given that the taxonomy used for annotating the presented corpus is at least to some extent unique, a direct comparison to models trained on other Slavic NER corpora is not straightforward as well. In particular, we do cover the classical PER, ORG and LOC types, however PRO and EVT are somewhat rare vis-a-vis other Slavic NER corpora reported in literature, also in terms of the definitions used. 


\subsection{Named-Entity Lemmatization}
\label{sec:lemmat-models}

In order to determine the baseline for the lemmatization task, we attempted three different approaches:

\begin{itemize}
    \item \textbf{orth} -- the lemma has the same value as the text form. This approach reflects the difficulty of the task, i.e. the amount of named entities that appear in an inflected form.
    \item \textbf{majority} -- the lemma is the most frequent lemma in the training corpus. If the text form is not present in the training subset, we take the text form as the lemma.
    \item \textbf{seq2seq} -- we trained a sequence-to-sequence neural network using the pre-trained multilingual mT5-large model \cite{xue2021mt5}.\footnote{\url{https://huggingface.co/google/mt5-large}} We use the following parameter settings for the training:
    \\[1em]
    \begin{tabular}{ll}
        Sequence length & 32 \\
        Epochs & 10 \\
        Learning rate & 2e-5 \\
        Optimizer & Adafactor \\
        Scheduler & Liner without warmup \\
        Weight decay & 0.01 \\
    \end{tabular}
    \end{itemize}

For the \textbf{cross topics} split, the baseline model achieved an accuracy of 96.13\% across all languages. The transformer-based approach outperforms the majority-based lemmatization by less than 4 pp. This is due to the fact that the cross-domain split has a relatively high overlap of entities between the training and testing subsets. The scores for each language are similar and vary between 94 and 97\%. The detailed results are presented in Table~\ref{tab:eval-cross-topics-languages-lemmatization}. 

\begin{table}[ht]
  \centering
  \begin{scriptsize}
    \begin{tabular}{@{}lrrrr}
      \toprule
  \textbf{Language}   & \textbf{Orth} & \textbf{Majority} & \textbf{Seq2seq} & \textbf{Support} \\ 
  \midrule
PL  &  51.82  & 93.40   & 96.68 & 2~986    \\
CS  &  56.92  & 93.80   & 97.83 & 1~567    \\
RU  &  50.30  & 89.73   & 94.09 & 3~458    \\ 
BG  &  82.17  & 95.05   & 97.45 & 2~872    \\
SL  &  57.54  & 88.30   & 94.38 & 1~976    \\
UK  &  54.09  & 95.00   & 98.36 & 1~100    \\
\midrule
All & 59.25  & 92.28    & 96.13 & 13~959\\
      \bottomrule
    \end{tabular}
    \end{scriptsize}
    \caption{Evaluation of the lemmatization on the \textbf{cross topics} split---accuracy for each language.}
  \label{tab:eval-cross-topics-languages-lemmatization}
\end{table}

For the \textbf{single-out topic} split, we can observe a significantly higher difference between the scores obtained by the majority-based approach and the transfer-based model -- the difference is more than 16 pp. The machine learning approach achieved an accuracy of 88.89\% across all languages. The range of scores for each language is higher and varies from 85 to 93\%. The detailed results are presented in Table~\ref{tab:eval-single-out-topic-languages-lemmatization}. 

\begin{table}[ht]
  \centering
  \begin{scriptsize}
    \begin{tabular}{@{}lrrrr}
      \toprule
  \textbf{Language}   & \textbf{Orth} & \textbf{Majority} & \textbf{Seq2seq} & \textbf{Support} \\ 
  \midrule
PL  &  59.16  &  77.41  &   90.13  &  2~549 \\
CS  &  54.70  &  74.67  &   90.06  &  1~137 \\
RU  &  48.79  &  68.05  &   86.60  & 18~018 \\ 
BG  &  76.61  &  82.81  &   88.51  &  6~085 \\
SL  &  52.87  &  79.20  &   93.45  &  7~082 \\
UK  &  49.23  &  55.77  &   91.07  & 3~085 \\
\midrule
All & 55.24 & 72.32 & 88.89 & 37~956 \\
      \bottomrule
    \end{tabular}
    \end{scriptsize}
    
    \caption{Evaluation of the lemmatization on the \textbf{single-out topic} split---accuracy for each language.}
  \label{tab:eval-single-out-topic-languages-lemmatization}
\end{table}


\subsection{Named-Entity Linking}
\label{sec:linking-models}

To establish the baseline for entity linking, we used the same approach as for named entity lemmatization with a sequence-to-sequence neural network. We used exactly the same configuration, except that the output was a cross-language identifier. The average accuracy for the \textbf{cross topics} split is 87.84\%, and the scores vary from 84 to 91\%. The evaluation results are provided in Table~\ref{tab:eval-cross-topics-languages-linking}.

\begin{table}[ht]
  \centering
  \begin{scriptsize}
    \begin{tabular}{@{}lrr}
      \toprule
  \textbf{Language}  & \textbf{Seq2seq} & \textbf{Support} \\ 
  \midrule
PL  & 89.92  & 2~986    \\
CS  & 88.13 & 1~567    \\
RU  & 85.13  & 3~458    \\ 
BG  & 89.58  & 2~872    \\
SL  & 84.86  & 1~976    \\
UK  & 91.09  & 1~100    \\
\midrule
All & 87.84 & 13~959\\
      \bottomrule
    \end{tabular}
    \end{scriptsize}
    \caption{Evaluation of the entity linking on the \textbf{cross topics} split---accuracy for each language.}
  \label{tab:eval-cross-topics-languages-linking}
\end{table}

For the \textbf{single-out topic} split, the average accuracy for all languages is 68.75\%, and the scores vary from 58 to 77\%. The evaluation results are provided in Table~\ref{tab:eval-single-out-topic-languages-linking}.

\begin{table}[ht]
  \centering
  \begin{scriptsize}
    \begin{tabular}{@{}lrrrr}
      \toprule
  \textbf{Language}   & \textbf{Seq2seq} & \textbf{Support} \\ 
  \midrule
PL  & 75.13  &   2~549  \\
CS  & 77.92  &   1~137  \\
RU  & 67.56  &  18~018  \\ 
BG  & 63.60   &  6~085 \\
SL  & 76.81   &  7~082  \\
UK  & 58.94  &   3~085 \\
\midrule
All & 68.75 & 37~956 \\
      \bottomrule
    \end{tabular}
    \end{scriptsize}
    \caption{Evaluation of the entity linking on the \textbf{single-out topic} split---accuracy for each language.}
  \label{tab:eval-single-out-topic-languages-linking}
\end{table}

It is important to mention that the accuracy figures of the baseline models provided here are not directly comparable with the ones used in the shared tasks~\cite{piskorski-etal-2019-second,piskorski-etal-2021-slav,yangarber-etal-2023-slav} since a different metric was used, namely, LEA~\cite{moosavi-strube-2016-coreference}.



\section{Access}

The corpus presented in this paper is publicly available for research purposes at \url{http://github.com/SlavicNLP/SlavicNER} and \url{https://bsnlp.cs.helsinki.fi/SlavicNER}. The baseline ML models are publicly available at \url{https://huggingface.co/SlavicNLP}. For any further use and questions related to the corpus please contact the authors of this paper.

\section{Conclusions}
\label{sec:conclusions}

This paper describes the construction of a multilingual resource for named entity recognition for Slavic languages.  NER is considered the fundamental information extraction (IE) task; it is also an essential sub-task in all higher-level IE tasks and in many real-world NLP applications, e.g.~\cite{piskorski-2013-IE-future,linge-2010-IE-medisys} 

This work results from a series of shared tasks conducted during 2017–2023 as part of the Workshops on Slavic Natural Language Processing.  The corpus comprises more than 5K documents on seven topics, and more than 152K entity mentions.  We provide a range of detailed statistics, including the distribution of documents, entities, and categories of entities across languages and topics, and the distribution of document lengths across languages. 

We provide two splits of the dataset for in-domain (cross-topic split) and cross-domain (single-topic-out) evaluation.  To calculate the baseline results for the splits, we trained two models for entity recognition and classification using the pre-trained multilingual \textit{XLM-RoBERTa-large} model.  The reported baseline performance is 92.22 and 85.03 micro-averaged F1-score, for the in-domain and cross-domain evaluations, respectively.

The resulting corpus serves as a highly adaptable resource, offering valuable utility for developing and evaluating models designed for a wide range of tasks, including NE recognition, categorization, lemmatization, and the establishment of co-reference connections among entity mentions.
This versatility extends to both mono-lingual and cross-lingual applications, making it an invaluable asset for research in the field.  The corpus is freely available for research purposes.  To our knowledge, this is the first and only multilingual Slavic Named Entity-related corpus with names being cross-lingualy linked and provided with base forms.  Other existing NER corpora for Slavic languages are strictly monolingual.  Furthermore, for the first time the integrated and curated corpus is available with positional anchoring.  We believe the corpus will contribute to fostering research not only on NER, but also name linking and name lemmatization for Slavic languages.

Future extensions of the corpus may include assigning more fine-grained labels to the entities labeled as PRO, since this is a rather broad category.  Analogously, the event category could be subdivided into more fine-grained labels, namely, occasions, incidents, natural disasters, phenomena, etc.

\section{Acknowledgements}

We are grateful to the following people for contributing to the annotation of the data or other activities which contributed to the creation of the presented dataset: Bogdan Babych, Anna Dmitrieva, Zara Kancheva, Olga Kanishcheva, Anisia Katinskaia, Laska Laskova, Maria Lebedeva, Preslav Nakov, Petya Osenova, Lidia Pivovarova, Senja Pollak,
Pavel Pribáň, Ivaylo Radev, Kiril Simov,  Marko Robnik-Šikonja, Piotr Rybak, Jan Šnajder, Vasyl Starko, and Josef Steinberger.

\section*{Ethics Statement}

We find no ethical issues with the current work.   We use publicly available resources for all conducted experiments, and release the language resources, which were created in collaboration with colleagues, who were aware of how the data will be used by the research community.

\section*{References} 

\bibliographystyle{lrec-coling2024-natbib}
\bibliography{lrec-coling2024-example,literature}

\appendix
\section*{Appendices}

\section{Annotation guidelines}
\label{ne-annotation-guidelines}

The Annotation concerns labeling five types of named entities: persons (PER), organizations (ORG), locations (LOC), events (EVT), and products (PRO).

Each named entity mention annotation should additionally include the lemma of the named entity and an identifier in such a way that detected mentions referring to the same real-world entity should be assigned the same identifier, which we will refer to as cross-lingual ID.

\subsection{General Rules}

\begin{itemize}

 \item When assigning the type to a named entity (ORG, LOC, PER, EVT or PRO) in general it is assumed that the local document context and common knowledge is considered and exploited for resolving ambiguities, e.g., in {\color{blue}``\textit{Twitter announced revenues for 2018 ...}''} the name {\color{blue}``\textit{Twitter}''} refers to a company (ORG), whereas in the phrase {\color{blue}``\textit{I posted it on Twitter}''} the name {\color{blue} ``\textit{Twitter}''} refers to a product (PRO), unless explicitly specified otherwise in the remaining part of these guidelines.

 \item In cases, in which the local document context and common knowledge does not provide sufficient information to disambiguate the named entity type (e.g., in the phrase {\color{blue}``\textit{Opel announced that ...}''} the mention of {\color{blue}``\textit{Opel}''} could potentially refer either to an organisation (ORG) or a person (PER), namely, Adam Opel, the founder of the company, the more probable interpretation should be considered, i.e., ORG, since Adam Opel died in 1837 and could not announce anything recently, unless the document is a historical one (unlikely). In case both NE type interpretation appear to be equally probable the following NE type disambiguation rules should be applied.
 If possible interpretation is either ORG or PER then PER should have priority. If possible interpretation is ORG or PRO then ORG should have priority.

 \item Sometimes entities are mentioned using a common noun starting with an uppercase letter, e.g., {\color{blue}``\textit{Senat}''} (Slovene: ``Senat''). If it is clear from the context that they refer to a specific entity (e.g. {\color{blue}``\textit{Senat Združenih držav Amerike}''} -- Slovene: ``United States Senate'') such mentions are to be treated as named mentions and should be tagged accordingly. Lowercase and lowercase-uppercase mixture mentions of the full name of an entity are to be treated as named mentions, e.g., {\color{blue}``\textit{vlada republike slovenije}''} (Slovene: ``The government of Repubic of Slovenia'') and {\color{blue}``\textit{vlada Republike Slovenije}''} should be considered as mentions of the entity with the full name {\color{blue}``\textit{Vlada Republike Slovenije}''}.

 \item Lemmatisation of a named entity mention refers to lemmatisation of the surface form extracted from the input text, e.g., the lemma for {\color{blue}``\textit{UE}''} and {\color{blue}``\textit{Unii Europejskiej}''} (Polish: genitive form of ``European Union'') is {\color{blue}``\textit{UE}''} and {\color{blue}``\textit{Unia Europejska}''} (Polish: nominative form of ``European Union'') respectively, whereas the cross-lingual ID assigned to the two aforementioned NE mentions should be the same. Analogously, lemma of a plural mention of an entity is expected to be nominative plural form (e.g., lemma of the word {\color{blue}``\textit{Japoncích}''} (Czech: genitive plural form of ``Japanese'') should be {\color{blue}``\textit{Japonci}''} -- Czech: nominative plural form of ``Japanese''), whereas lemma of a singular mention should be nominative singular (e.g., lemma of the word {\color{blue}``\textit{Japoncem}''} (Czech: genitive singular ``Japanese'') should be {\color{blue}``\textit{Japonec}''}).

 \end{itemize}

 \subsection{Persons (PER)}
 \label{sec:per}

 \begin{enumerate}

 \item This category covers named references to individual people (e.g., {\color{blue}``\textit{Donald Trump}''}) or families (e.g. {\color{blue}``\textit{Kaczyńscy}''}), and certain named references to groups of people.

 \item Person names should not include titles, honorifics, and functions/positions. For example, in the text fragment {\color{blue}``\textit{CEO Dr. Jan Kowalski}''}, only {\color{blue}``\textit{Jan Kowalski}''} should be annotated as a person name. However, initials and pseudonyms are considered named mentions of person names and should be annotated. Analogously, in the text fragment {\color{blue}``\textit{The Prime Minister of the United Kingdom Theresa May}''}, only {\color{blue}``\textit{Theresa May}''} should be tagged as PER, whereas {\color{blue}``\textit{United Kingdom}''} should be tagged as LOC.

 \item Personal possessives derived from a named mention of a person should be annotated and classified as a person. For instance, for {\color{blue}``\textit{Trumpov tweet}''} (Croatian - {\color{blue}``\textit{Trump’s email}''}) it is expected to  annotate {\color{blue}``Trumpov''} as PER and have as the base form: {\color{blue}``\textit{Trump}''}.

 \item \label{item:per-top} Toponym-based (e.g., country name adjectives) references to groups of people that are linked to geopolitical entities\footnote{Geo-Political Entities are considered to be complex entities consisting of a population, a government or some administrative body, and a physical location. In the context of this shared task geo-political entities comprise countries, provinces, states, counties and cities, and suchlike entities. In the context of this task international bodies and organisations are not considered geopolitical entities.
 } should also be annotated and tagged as PER, e.g., {\color{blue}``\textit{Ukrainians}''}. In this context, mentions of a single member belonging to such groups, e.g., {\color{blue}``\textit{Ukrainian}''} should be assigned the same cross-lingual ID as plural mentions, i.e., {\color{blue}``\textit{Ukrainians}''}. Furthermore, it should not matter whether {\color{blue}``\textit{Ukrainians''}} refer to the entire nation, some “unspecified” part thereof or an organisation related to the geopolitical entity. In all these cases PER category should be used and the cross-lingual ID should be the same as assigned to other toponym-based references to the same geopolitical entity, e.g., {\color{blue}``\textit{Ukrainians}''} and {\color{blue}``\textit{Ukraine}''} should be assigned the same cross-lingual ID (different mention type, but the same cross-lingual ID).

 \item Although continents ({\color{blue}``\textit{Europe}''}) and geographical regions ({\color{blue}``\textit{Eastern Europe}''}) are not geo-political entities, as regards named references to groups of people derived from such toponyms the same rule as specified in rule~\ref{item:per-top} applies by analogy, e.g., {\color{blue}``\textit{Europejczycy}''} (Polish: ``Europeans'') and {\color{blue}``\textit{Europa}''} (Polish: ``Europe'') should be classified as PER and LOC respectively, and both should be assigned the same cross-lingual ID.

 \item Named mentions of other groups of people that do have a formal organization unifying them should be tagged as PER and associated with the same cross-lingual ID as the mentions of the corresponding organisation, e.g., in the phrase {\color{blue}``\textit{Sparťané vyhráli...}''} (Czech: ``Spartans won...'') the mention of {\color{blue}``\textit{Sparťané}''} should be tagged as PER and have the same cross-lingual ID as the corresponding sport team, e.g., {\color{blue}``\textit{AC Sparta Praha}''} (football club). Analogously, phrases like {\color{blue}``\textit{Europoslanci}''} (Czech: ``Members of the European Parliament'') should be cross-linked with the mentions of the European Parliament (ORG).

 \item Mentions to groups of people that do not have a formal organization unifying them should not be annotated, e.g., phrases like {\color{blue}``\textit{Muslims}''}, which refers to a religious group not linked to any particular organisation should not be tagged.

 \item Fictive persons and characters (e.g., {\color{blue}``\textit{James Bond}'')} are considered as persons.

 \end{enumerate}

 \subsection{Locations (LOC)}
 \label{sec:loc}

 \begin{enumerate}

 \item This category includes all toponyms (e.g., cities, counties, provinces, regions, bodies of water, geological formations, etc.) and named mentions of facilities, i.e., functional and  primarily man-made structures, such as: stadiums, parks, museums, theaters, hotels, hospitals, transportation hubs (e.g. airports, sea ports, train stations), churches, streets, railroads, highways, bridges, tunnels, parkings, and other similar urban and non-urban facilities. For instance, {\color{blue}``\textit{Łazienki Królewskie w Warszawie}''} (Polish: ``Łazienki Park in Warsaw'') should be tagged as LOC, whereas general references to facilities without a concrete location as {\color{blue}``\textit{parki w Warszawie}''} (Polish: ``parks in Warsaw'') should not be annotated, i.e, only {\color{blue}``\textit{Warszawie}''} (Polish: locative form of ``Warsaw'') should be annotated as LOC in the latter case.

 \item \label{item:second} Even in case of named mentions of facilities that refer to an organization, the LOC tag should be used. For example, in the text {\color{blue}``\textit{The Schipol airport has acquired new electronic gates}''} the mention {\color{blue}``\textit{The Schipol airport}''} should be annotated and classified as LOC. 

 \item By analogy to rule~\ref{item:second} in~\ref{sec:loc}, toponyms, in particular, country names (e.g., {\color{blue}``\textit{Polska}''}) that refer to geopolitical entities (physical location, population of the country, respective government, nation, or a sport team representing a country) should be annotated and classified as LOC disregarding the specific named mention role ({\color{blue}``\textit{Poland}''} refers to a range of concepts) and assigned the same unique cross-lingual ID. In this context, the relevant toponyms and toponym-derived adjectives (see the rule~\ref{item:per-top} in Section~\ref{sec:per}) referring to the same geo-political entity should be assigned the same cross-lingual ID. In all other contexts, i.e., when a country name (or other toponym) is used to refer to an organisation that has no specific link to the respective geopolitical entity (unless it is accidental), e.g. a band named {\color{blue}``\textit{Russia}''}, it should be tagged as ORG.

 \item When recognising named mentions of facilities potential mentions of the location are considered to be part of the full mention, e.g., the entire phrase {\color{blue}``\textit{St. Stephen Church in Istanbul}''} should be annotated and tagged as LOC.

 \end{enumerate}

 \subsection{Organizations (ORG)}

 \begin{enumerate}

 \item This category covers all kind of organizations such as: political parties, public institutions, government units, non-governmental organizations, international organizations (e.g., European Union, NATO, united Nations, etc.) military organizations, companies, religious organizations, sport teams and organizations, education and research institutions, music groups, entertainment and media organizations, etc.

 \item Organization designators and potential mentions of the seat of the organization are considered to be part of the organization name. For instance, in the text fragment {\color{blue}``\textit{Citi Handlowy w Poznaniu}''} (Polish: ``City Handlowy bank in Poznań''), the full phrase {\color{blue}``\textit{Citi Handlowy w Poznaniu}''} should be annotated.
    
 \end{enumerate}

 \subsection{Events (EVT)}

 \begin{enumerate}

 \item This category covers named mentions of events, including: (a) occasions such as conferences, e.g. {\color{blue}``\textit{24. Konference Žárovného Zinkování}''}, concerts, festivales,  holidays, e.g., {\color{blue}``\textit{Święta Bożego Narodzenia}''} (Polish: ``Christmas''), (b) incidents such as wars, battles, and man-made disasters, e.g., {\color{blue}``\textit{Katastrofa Czernobylska}''} (Polish: ``Chernobyl catastrophe''), and (c) natural disasters and phenomena (including for instance: earthquakes, e.g. {\color{blue}``\textit{Great Alaska Earthquake}''}, volcano eruptions {\color{blue}``\textit{Eyjafjallajökull Eruption}''}, outbreaks of infectious diseases, e.g. {\color{blue}``\textit{Spanish Flu}''} , {\color{blue}``\textit{ebola}''} etc.).

 \item Future, speculative and fictive events, e.g., {\color{blue}``\textit{Czexit}''} or {\color{blue}``\textit{Polexit}''} are considered as event mentions too.

 \item In case a named mention of the event does also refer to a location, then it should be tagged as LOC. For example, in the phrase {\color{blue}``\textit{He died in Waterloo, just before the end of the battle}''}, {\color{blue}``\textit{Waterloo}''} should be tagged as LOC, not as EVT. However, the cross-lingual ID assigned to such a mention, i.e., {\color{blue}``\textit{Waterloo}''} should be the same as in the case of other mentions to the battle, e.g., {\color{blue}``\textit{The Battle of Waterloo}''}.

 \item When annotating named mentions of events the potential mentions of the location are considered to be part of the full mention, e.g., the entire phrase {\color{blue}``\textit{2004 Winter Olympics in Canada}''} should be annotated as EVT.
    
 \end{enumerate}

 \subsection{Products (PRO)}

 \begin{enumerate}

 \item This category covers product names, including for instance: electronics (e.g., {\color{blue}``\textit{Motorola Moto Z Play}''}, cars (e.g. {\color{blue}``\textit{Subaru Forester XT}''}), vehicles (e.g., {\color{blue}``\textit{Fiat Panda}''}), weapons (e.g., {\color{blue}``\textit{Kalashnikov AK-47}''}), web-based services (e.g., {\color{blue}``\textit{Twitter}''}), medicines (e.g., {\color{blue}``\textit{Oxycodone}''}, {\color{blue}``\textit{remdesivir}''}), stock (e.g., {\color{blue}``\textit{Google stock}''}), awards (e.g., {\color{blue}``\textit{Nobel Prize}''}), books (e.g., {\color{blue}``\textit{Harry Potter and the Sorcerer's Stone}''}), software (e.g., {\color{blue}``\textit{MS Office}''}), films (e.g., {\color{blue}``\textit{Gone with the Wind}''}), TV programmes (e.g., {\color{blue}``\textit{Wiadomości TVP}''}), newspapers (e.g., {\color{blue}``\textit{The New York Times}''}) and other pieces of art, etc.

 \item Names of legal documents, e.g., {\color{blue}``\textit{dyrektywy 2001/14/we Parlamentu Europejskiego i Rady}''} (Polish: ``directives 2001/14/we of the European Parliament and Council''), treaties, e.g., {\color{blue}``\textit{Traktat Lizboński}''} (Polish: ``Treaty of Lisbon''), initiatives/programmes (e.g., {\color{blue}``\textit{Horizon 2020}''}) are also considered product names

 \item  When a company name is used to refer to a {\em service}, e.g., {\color{blue}``\textit{na Instagramie}''} (Polish for ``\textit{on Instagram}''), the mention of {\color{blue}``\textit{Instagramie}''} is considered to refer to a service/product and should be tagged as \textsc{PRO}.  However, when a company name refers to a service expressing an opinion of the company, it should be tagged as \textsc{ORG}.

\end{enumerate}

 \subsection{Complex names}

 \begin{enumerate}

 \item In case of complex named entities, consisting of nested named entities, only the top-most (longest) entity should be annotated. For example, in the text fragment {\color{blue}``\textit{George Washington University}''} one should not annotate {\color{blue}``\textit{George Washington}''}, but the entire name, namely, {\color{blue}``\textit{George Washington University}''}.

 \item In case of coordinated phrases like for instance {\color{blue}``\textit{European and British Parliament}''} two names should be annotated (as ORG), i.e., {\color{blue}``\textit{European}''} and {\color{blue}``\textit{British Parliament}''}. The respective lemmas would be {\color{blue}``\textit{European}''} and {\color{blue}``\textit{British Parliament}''}. Furthermore, the IDs assigned to these entity mentions should refer to ``\textit{European Parliament}'' and ``\textit{British Parliament}'' respectively.

 \item In rare cases, plural forms might have two annotations---e.g., in the phrase {\color{blue}``\textit{a border between Irelands}''}---{\color{blue}``\textit{Irelands}''} should be annotated twice with identical lemmas but different IDs.

 \end{enumerate}

\newpage
\section{Detailed Statistics}
\label{sec:stats}

\begin{table}[h]
   \centering
   \begin{scriptsize}
     \begin{tabular}{@{} lrrrrrr}
       \toprule
   & \multicolumn{6}{c}{\em Asia Bibi}\\    
 lang & {\sc PL} & {\sc CS} & {\sc RU} & {\sc BG} & {\sc SL} & {\sc UK}\\
 \midrule
 {\sc PER} & 963 & 790 & 985 & 73 & 47 & 51 \\
 {\sc LOC} & 553 & 447 & 764 & 623 & 29 & 67\\
 {\sc ORG} & 354 & 264 & 494 & 280 & 10 & 38\\
 {\sc PRO} & 62 & 48 & 50 & 73 & 3 & 1\\
 {\sc EVT} & 14 & 3 & 1 & 13 & 0 & 0\\
       \midrule
       Total & 1946 & 1552 & 2294 & 2042 & 89 & 157\\
       \midrule
       forms & 508 & 303 & 407 & 412 & 51 & 87\\
       lemmas & 412 & 248 & 317 & 360 & 41 & 77\\
       entity IDs & 273 & 160 & 178 & 230 & 31 & 64\\
       \bottomrule
     \end{tabular}
     \end{scriptsize}
     \caption{Distribution of named-entity types across languages and statistics on surface forms, lemmas, and unique entities IDs for the domain {\em Asia Bibi}.}
   \label{tab:type-breakdown-asia}
 \end{table}

 \begin{table}[h]
   \centering
   \begin{scriptsize}
     \begin{tabular}{@{} lrrrrrr}
       \toprule
   & \multicolumn{6}{c}{\em Brexit}\\    
 lang & {\sc PL} & {\sc CS} & {\sc RU} & {\sc BG} & {\sc SL} & {\sc UK}\\
 \midrule
 {\sc PER} & 3607 & 1375 & 1795 & 4003 & 720 & 357 \\
 {\sc LOC} & 5269 & 1678 & 906 & 5151 & 564 & 643 \\
 {\sc ORG} & 5430 & 1604 & 1096 & 4830 & 542 & 203 \\
 {\sc PRO} & 906 & 293 & 153 & 633 & 32 & 18 \\
 {\sc EVT} & 1702 & 756 & 519 & 2277 & 414 & 252\\
       \midrule
       Total & 16914 & 5706 & 4481 & 16894 & 2287 & 1473 \\
       \midrule
       forms & 2820 & 1112 & 782 & 1212 & 596 & 234 \\
       lemmas & 2133 & 841 & 568 & 1103 & 411 & 177 \\
       entity IDs & 1507 & 582 & 268 & 781 & 287 & 127 \\
       \bottomrule
     \end{tabular}
     \end{scriptsize}
     \caption{Distribution of named-entity types across languages and statistics on surface forms, lemmas, and unique entities IDs for the domain {\em Brexit}.}
   \label{tab:type-breakdown-brexit}
 \end{table}

 \begin{table}[h]
   \centering
   \begin{scriptsize}
     \begin{tabular}{@{} lrrrrrr}
       \toprule
   & \multicolumn{6}{c}{\em Nord Stream}\\    
 lang & {\sc PL} & {\sc CS} & {\sc RU} & {\sc BG} & {\sc SL} & {\sc UK}\\
 \midrule
 {\sc PER} & 692 & 681 & 454 & 404 & 814 & 87 \\
 {\sc LOC} & 2107 & 2174 & 1885 & 1574 & 2110 & 655 \\
 {\sc ORG} & 1066 & 638 & 1115 & 747 & 834 & 777 \\
 {\sc PRO} & 1150 & 616 & 750 & 654 & 370 & 8 \\
 {\sc EVT} & 17 & 19 & 5 & 8 & 57 & 17 \\
       \midrule
       Total & 5032 & 4129 & 4210 & 3387 & 4185 & 1544\\
       \midrule
       forms & 845 & 770 & 892 & 504 & 902 & 336\\
       lemmas & 634 & 550 & 583 & 448 & 600 & 244\\
       entity IDs & 441 & 392 & 320 & 305 & 461 & 175\\
       \bottomrule
     \end{tabular}
     \end{scriptsize}
     \caption{Distribution of named-entity types across languages and statistics on surface forms, lemmas, and unique entities IDs for the domain {\em Nord Stream}.}
   \label{tab:type-breakdown-nord}
 \end{table}

 \begin{table}[h]
   \centering
   \begin{scriptsize}
     \begin{tabular}{@{} lrrrrrr}
       \toprule
   & \multicolumn{6}{c}{\em Ryanair}\\    
 lang & {\sc PL} & {\sc CS} & {\sc RU} & {\sc BG} & {\sc SL} & {\sc UK}\\
 \midrule
 {\sc PER} & 163 & 171 & 74 & 194 & 151 & 36 \\
 {\sc LOC} & 1112 & 1071 & 1092 & 526 & 584 & 618 \\
 {\sc ORG} & 967 & 1170 & 1057 & 359 & 662 & 520 \\
 {\sc PRO} & 136 & 77 & 83 & 90 & 148 & 20 \\
 {\sc EVT} & 8 & 13 & 0 & 7 & 10 & 0 \\
       \midrule
       Total & 2386 & 2502 & 2306 & 1176 & 1558 & 1194\\
       \midrule
       forms & 514 & 475 & 400 & 323 & 671 & 187 \\
       lemmas & 418 & 400 & 332 & 315 & 519 & 137 \\
       entity IDs & 322 & 306 & 251 & 245 & 426  & 108\\
       \bottomrule
     \end{tabular}
     \end{scriptsize}
     \caption{Distribution of named-entity types across languages and statistics on surface forms, lemmas, and unique entities IDs for the domain  {\em Ryanair}.}
   \label{tab:type-breakdown-ryanair}
 \end{table}

 \begin{table}[h]
   \centering
   \begin{scriptsize}
     \begin{tabular}{@{} lrrrrrr}
       \toprule
   & \multicolumn{6}{c}{\em Covid}\\    
 lang & {\sc PL} & {\sc CS} & {\sc RU} & {\sc BG} & {\sc SL} & {\sc UK}\\
 \midrule
 {\sc PER} & 552 & 583 & 754 & 489 & 1098 & 268 \\
 {\sc LOC} & 507 & 579 & 1184 & 1265 & 1583 & 609 \\
 {\sc ORG} & 662 & 399 & 910 & 909 & 1379 & 758 \\
 {\sc PRO} & 175 & 194 & 744 & 211 & 373 & 187 \\
 {\sc EVT} & 376 & 502 & 1309 & 806 & 1058 & 567 \\
       \midrule
       Total & 2272 & 2257 & 4907 & 3680 & 5503 & 2389 \\
       \midrule
       forms & 688 & 941 & 1436 & 1111 & 2191 & 625 \\
       lemmas & 557 & 745 & 1128 & 1016 & 1770 & 509 \\
       entity IDs & 404 & 558 & 793 & 764 & 1393 & 369 \\
       \bottomrule
     \end{tabular}
     \end{scriptsize}
     \caption{Distribution of named-entity types across languages and statistics on surface forms, lemmas, and unique entities IDs for the domain {\em Covid}.}
   \label{tab:type-breakdown-covid}
 \end{table}

 \begin{table}[h]
   \centering
   \begin{scriptsize}
     \begin{tabular}{@{} lrrrrrr}
       \toprule
   & \multicolumn{6}{c}{\em US Elections 2020}\\    
 lang & {\sc PL} & {\sc CS} & {\sc RU} & {\sc BG} & {\sc SL} & {\sc UK}\\
 \midrule
 {\sc PER} & 881 & 553 & 6214 & 3141 & 4001 & 1136 \\
 {\sc LOC} & 1091 & 340 & 6324 & 1699 & 1879 & 892 \\
 {\sc ORG} & 299 & 118 & 3852 & 762 & 743 & 491 \\
 {\sc PRO} & 112 & 69 & 1138 & 296 & 325 & 137 \\
 {\sc EVT} & 110 & 66 & 486 & 190 & 137 & 435 \\
       \midrule
       Total & 2550 & 1137 & 18065 & 6088 & 7089 & 3091\\
       \midrule
       forms & 475 & 378 & 3614 & 1124 & 1606 & 541 \\
       lemmas & 349 & 279 & 2650 & 1019 & 1129 & 390 \\
       entity IDs & 270 & 201 & 1700 & 667 & 833 & 270 \\
       \bottomrule
     \end{tabular}
     \end{scriptsize}
     \caption{Distribution of named-entity types across languages and statistics on surface forms, lemmas, and unique entities IDs for the domain {\em US Elections 2020}.}
   \label{tab:type-breakdown-us}
 \end{table}

 \begin{table}[h]
   \centering
     \begin{tabular}{@{} lrrr}
       \toprule
   & \multicolumn{3}{c}{\em Russia-Ukraine War}\\    
 lang & {\sc PL} & {\sc CS} & {\sc RU}\\
 \midrule
 {\sc PER} & 399 & 277 & 279 \\
 {\sc LOC} & 964 & 459 & 700 \\
 {\sc ORG} & 302 & 196 & 436 \\
 {\sc PRO} & 99 & 59 & 107 \\
 {\sc EVT} & 68 & 54 & 17 \\
       \midrule
       Total & 1832 & 1045 & 1539 \\
       \midrule
       forms & 723 & 498 & 725 \\
       lemmas & 563 & 384 & 594 \\
       entity IDs & 410 & 280 & 493 \\
       \bottomrule
     \end{tabular}
     \caption{Distribution of named-entity types across languages and statistics on surface forms, lemmas, and unique entities IDs for the domain {\em Russia-Ukraine War}.}
   \label{tab:type-breakdown-ru-ukr-war}
  \end{table}

\end{document}